# DataLearner: A Data Mining and Knowledge Discovery Tool for Android Smartphones and Tablets

Darren Yates[1], Md Zahidul Islam[1] and Junbin Gao[2]

[1] Charles Sturt University, Panorama Ave, Bathurst NSW 2795, Australia
[2] University of Sydney Business School, The University of Sydney, NSW, 2006, Australia
[1]`{dyates, zislam}@csu.edu.au`
[2]`junbin.gao@sydney.edu.au`

**Abstract.** Smartphones have become the ultimate 'personal' computer, yet despite this, general-purpose data mining and knowledge discovery tools for mobile devices are surprisingly rare. DataLearner is a new data mining application designed specifically for Android devices that imports the Weka data mining engine and augments it with algorithms developed by Charles Sturt University. Moreover, DataLearner can be expanded with additional algorithms. Combined, DataLearner delivers 40 classification, clustering and association rule mining algorithms for model training and evaluation without need for cloud computing resources or network connectivity. It provides the same classification accuracy as PCs and laptops, while doing so with acceptable processing speed and consuming negligible battery life. With its ability to provide easy-to-use data mining on a phone-size screen, DataLearner is a new portable, self-contained data mining tool for remote, personalised and educational applications alike. DataLearner features four elements – this paper, the app available on Google Play, the GPL3-licensed source code on GitHub and a short video on YouTube.

## 1 Introduction

The growing demand for knowledge discovery and data mining has resulted in numerous software tools being developed for PCs and laptops, including the likes of Weka [1] and RStudio [2]. These tools enable users to execute an array of classification, clustering and association rule mining on datasets of the user's choosing.

Meanwhile, since their introduction, smartphones have become the ultimate version of the 'personal' computer, providing multi-core processing power, memory and storage inside a compact battery-powered device. In addition, today's smartphones and tablets are equipped with an array of sensors, including accelerometers and gyroscopes for movement, microphone and speaker for audio, temperature, humidity and barometric pressure sensors for environment. Yet, the one device class where general-purpose knowledge discovery and data mining tools remain rare is mobile devices.

DataLearner is a new open-source data mining and knowledge discovery tool designed for smartphones and tablets running the Android operating system, beginning with Android 4.4 (codenamed 'KitKat'). It includes 40 classification, clustering and association rule mining algorithms and features separate automatically-assigned user interfaces for smartphone and tablet use. DataLearner is fully self-contained – no cloud computing resources are required to train or 'build' models and no external storage devices are needed. This presents numerous advantages. First, as no cloud computing or networking connectivity is required, DataLearner enables a smartphone to become a portable data mining tool in remote-location applications where mains power or network connectivity may not be available. Second, by being fully self-contained, the app does not require any data be sent to or received from a server in order to build a model, thus significantly improving data security and privacy. Third, by being able to locally build a model from local data, there is greater scope for personalised feedback, particularly in applications such as personal e-health. Finally, it also provides a practical and compact learning alternative for the growing numbers of students globally studying data science, allowing them to augment their PCs and laptops and build models directly on their Android mobile devices.

### 1.1 Our Contributions

This paper details the process of developing a mobile application capable of locally-executed data mining and thus, includes a number of contributions:

- To our knowledge, DataLearner is the only application of its type on Google Play, filling a void within the mobile software market.
- DataLearner can be expanded through the inclusion of additional algorithms. Examples of added algorithms are detailed in Sections 3.3 and 3.4.
- DataLearner overcomes issues restricting the initial use of the Weka open-source data mining core within the Android platform. These are detailed in Section 3.5.
- DataLearner features a user interface designed to enable easy model-building and evaluation even on a small 800x480-pixel phone screen.
- The DataLearner platform consists of four elements – this research paper, the GPL3-licenced source code available on Github[1], a short video tutorial on YouTube[2] and the DataLearner application available on Google Play[3].

Our vision for DataLearner is to enable mobile data mining and knowledge discovery in applications ranging from remote-location processing to education. Thus, just as the Weka data mining suite focused on the end user [3], so too DataLearner aims to support the end user in need of a compact and easy-to-use data mining tool, but who may have swapped a PC or laptop for an Android smartphone or tablet.

The remainder of this paper continues with Section 2 backgrounding previous research efforts in mobile data mining. Section 3 details the techniques and methods behind DataLearner, with Section 4 outlining basic usage. Section 5 covers the performance of DataLearner in terms of accuracy and speed on smartphone hardware compared with a desktop PC using a range of datasets and algorithms. Section 6 looks at future improvements and research efforts, while Section 7 concludes this paper.

---

[1] https://github.com/darrenyatesau/DataLearner
[2] https://youtu.be/H-7pETJZf-g
[3] https://play.google.com/store/apps/details?id=au.com.darrenyates.datalearner

## 2 Related Work

Since their arrival in 2007, smartphones have progressed from mobile phones with basic apps to personal computers with built-in sensors, 3D graphics engines and multiple wireless communications. This led to early attempts to incorporate local data mining onto a mobile device. For example, the 'Mobileminer' application [4] provided a 'general purpose service' for the task of mining frequent patterns. In addition, the 'acquisitional context engine' (ACE) was 'middleware' code used in identifying user activities with mobile devices through association rule mining [5]. However, Mobileminer was designed for the Samsung Tizen operating system, while ACE operated on Windows Mobile 7.5. Thus, neither application is natively operable on the Android platform. Today, Google's TensorFlow Lite [6] now offers reduced-scale neural network processing on mobile devices, but research into more general-purpose data mining and knowledge discovery methods on Android devices remains limited.

Weka is a popular graphical user interface (GUI)-based data mining application developed by the University of Waikato to support Windows, Linux and macOS operating systems [7]. It enables the user to build models from training datasets using a range of classification, clustering and association rule algorithms. As such, it has become a popular choice in learning institutions, allowing both experienced and inexperienced users to achieve useful results. However, despite sharing a code base with Linux, Android is not specifically listed as a supported operating system for Weka. Nevertheless, being open-source, Weka features access to its core engine through a Java archive (.jar) file. Previous attempts to bring Weka to Android do exist, including health-focused mHealthDroid [8] and Weka-for-Android [9], but both appear to have been deprecated, with neither receiving source code updates since at least 2014.

Moreover, as of June 2019, there appear to be no applications on the Google Play store enabling easy-to-use data mining and knowledge discovery on Android devices. With smartphones being increasingly employed in data-gathering roles varying from mental health [10, 11] to road maintenance [12], there is significant potential for smartphones to evolve into self-contained, portable data mining tools that do not rely on cloud computing or network connectivity. The result is a notable gap not just in terms of research, but also in the application development of mobile data mining.

Previous research we have conducted has found that the Weka core engine is operable within the Android platform [13]. That research revealed through early testing that Android smartphones present no issues in terms of classification accuracy, delivering the same accuracy performance as personal computers (PCs). Moreover, additional testing found the effect of locally executing data mining algorithms on smartphone battery life is generally negligible. Thus, a general-purpose locally-executed data mining application should be capable of running on Android devices. This paper will now detail the theory and implementation of such an application we have called 'DataLearner'.

## 3 Our Design Goals and Implementation

Our concept was to develop a general-purpose Android application that would allow any user, without requiring programming experience, to load a Weka-compatible dataset, select a suitable algorithm, build a model and test the accuracy of that model using 10-fold cross-validation on any available Android-compatible device. A further goal was to enable the application to expand through the addition of new algorithms as required. This section details these technical and design decisions in creating the DataLearner application.

### 3.1 User interface design

Two of the design goals set for the user interface were ease-of-use by novice data-miners as well as ease-of-access to all features on a phone-sized screen. These were achieved by breaking down the application's functions into three key areas designated 'load', 'select' and 'run' – load a training dataset, select an algorithm, run the algorithm. Each key area is split into a self-contained Android window or 'fragment', accessible by either swipe-left and right movements on the screen, or by pressing the appropriate top menu tab option. Moreover, human-computer interaction (HCI) research has shown that the addition of horizontal swiping has a positive effect on a user's intentions to use mobile content [14]. These fragments are shown in Fig. 1.

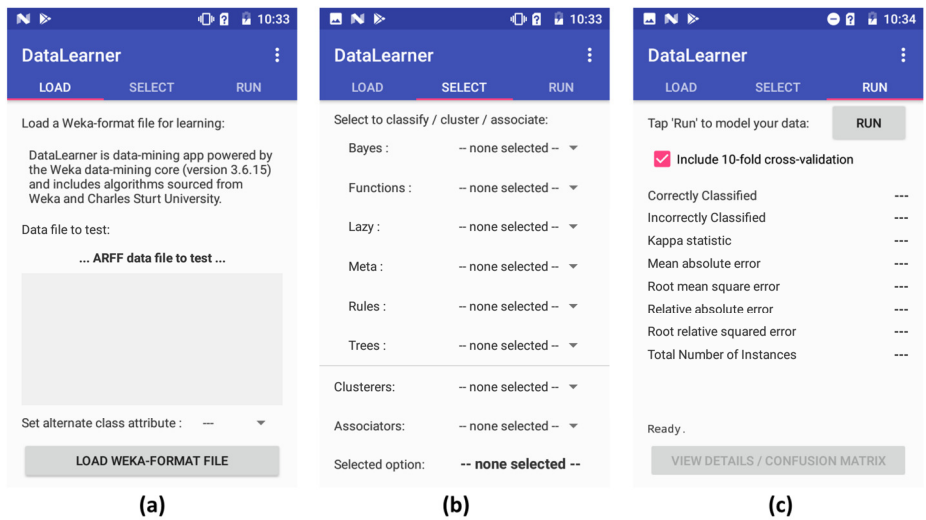

**Fig. 1.** DataLearner's three user interface screens - (a) Load, (b) Select and (c) Run.

### 3.2 Algorithm selection

As the Weka core engine provides a significant number of built-in algorithms, it was decided to implement as many of these as practical within the DataLearner applica-tion. In the first release of DataLearner, this amounts to 37 Weka-based classification, clustering and association rule mining algorithms, with the classifiers divided into six subgroups as follows:

- **Bayes** [15] – BayesNet, NaiveBayes
- **Functions** [16] – Logistic, SimpleLogistic
- **Lazy** [17] – IBk (K Nearest Neighbours), KStar
- **Meta** [18] – AdaBoostM1, Bagging, LogitBoost, MultiBoostAB, Random Committee, RotationForest
- **Rules [19]** – Conjunctive Rule, Decision Table, DTNB, JRip, OneR, PART, Ridor, ZeroR
- **Trees** [20] – ADTree, BFTree, DecisionStump, J48 (C4.5), LADTree, Random Forest, RandomTree, REPTree, SimpleCART.

The clusterers [21] include DBSCAN, Expectation Maximisation (EM), FarthestFirst, FilteredClusterer and SimpleKMeans, while the Association Rule algorithms [22] feature Apriori, FilteredAssociator and FPGrowth.

### 3.3    Adding external algorithms

In addition to the collection of built-in algorithms discussed in Section 3.2, the DataLearner source code has been designed to also allow external algorithms to be included and compiled into the application. This process has enabled the addition of three further classifiers developed by researchers at Charles Sturt University, namely SysFor [23], ForestPA [24] and SPAARC [25].

This is achieved by replicating the standard Weka source folder structure within the DataLearner Java application folder. For example, the common path for storing decision tree algorithms within Weka is /weka/classifiers/trees. By the addition of a 'weka.classifiers.trees' Android package to the source code, new decision tree algorithms can be included. As this initial release of DataLearner uses version 3.6.15 of the Weka core engine, any suitable algorithm conforming to this version should be accessible by the Weka core and execute normally (this includes the cross-validation process). This was the technique used to successfully incorporate the SysFor, ForestPA and SPAARC classification algorithms.

### 3.4    SysFor, ForestPA and SPAARC

These three classification algorithms, developed at Charles Sturt University, were chosen to show the expandability of DataLearner, but also to provide additional capabilities well suited for mobile devices.

SysFor [23] is an ensemble classifier capable of building a forest of decision trees from both high and low-dimensional (attribute count) datasets. It uses a systematic approach to building trees based on the 'goodness' of attributes (those that achieve higher gain ratios during node testing), even when the dimensionality is low compared to the desired number of trees. SysFor's ability to obtain good results from low-dimensional datasets in particular makes it well suited for mobile devices, where datasets of low size and dimension are likely to be more common.

ForestPA [24] is another ensemble classifier but takes a different approach. It aims to overcome a perceived limitation of RandomForest that can lead to uneven selection of good- and poor-performing attributes during node testing. ForestPA generates trees in sequence, testing all attributes at each node, not just a subset. To create diversity amongst the trees, ForestPA adds a penalty to an attribute appearing in the latest tree, the closer the attribute to the root node, the greater the penalty. On completion of each tree, the weights are adjusted, allowing older attributes the opportunity to reappear in future trees. The final result is an improvement in classification over RandomForest, particularly where a training dataset may only have a few good-quality attributes.

SPAARC (Split-Point and Attribute-Reduction Classifier) [25] is a single-tree classification algorithm based on the popular CART (Classification And Regression Tree) algorithm [26]. It incorporates techniques for reducing the computational load, improving processing time whilst minimising effects on classification accuracy. These include 'split-point sampling' that reduces the number of split-points used when testing the suitability of an attribute at each node in the tree, plus 'node-attribute sampling', whereby a subset of attributes is tested at each alternate horizontal tree node level. Tests revealed the combined effect over a number of training datasets is the reduction in model-build time of as much as 69%. This reduced run-time is an ideal feature well suited to constrained devices such as smartphones.

### 3.5  Solving integration issues

In order to achieve a self-contained application, the Weka run-time core has to be imported into the Android Studio development environment. However, this revealed some issues during development. First, it was found that the latest stable version 3.8 release of Weka appears to be incompatible with Android. This was due to a change implemented in Weka 3.7 and following that incorporates a user interface based on Java's Swing/Abstract Window Toolkit (AWT), which is not fully supported by Android. Further investigation revealed that reverting to version 3.6, which implements a standard Java interface, enabled successful operation. Thus, this initial release of our DataLearner application features the latest 3.6.15 release of the Weka core engine.

In addition to the Weka/AWT problem, a separate issue was identified involving J48, Weka's implementation of the C4.5 decision tree algorithm. This issue resulted in a failure to successfully build and evaluate models from some datasets with more than approximately 5,000 records. Further troubleshooting identified the issue involved the Android 'stack', a general-purpose Java memory block for holding active variable data. Since Android was initially designed to operate on hardware-constrained devices, the stack size in early Android releases was set to a low level of 8KB or less. However, by dynamically adjusting the stack to 64KB, the J48 algorithm now successfully completes the model build/evaluation process with the same datasets that originally failed to. No further J48 issues have since been detected.

DataLearner requires a device running a minimum of Android 4.4 (codenamed 'KitKat'). Supporting this minimum enables DataLearner to cater for 96.2% of all active devices accessing the Google Play store repository during the week ending May 7, 2019 [27]. At this point, none of the algorithms have been modified to en-hance performance on mobile devices. Moreover, no limits have been introduced on dataset size or dimensionality. However, overall processing time will be a function of dataset size, the algorithm selected and the processing speed of the mobile device.

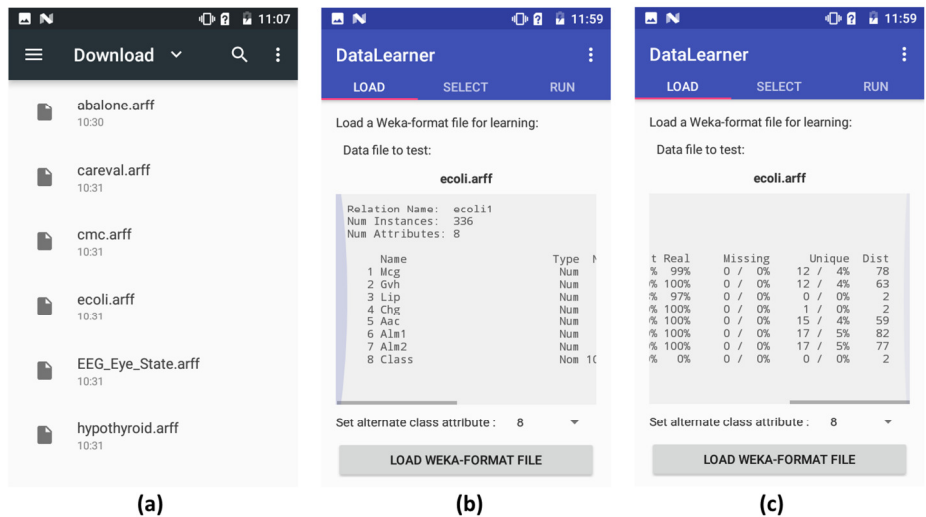

**Fig. 2.** Once an ARFF-format dataset is selected (a), a summary appears in the scrollable summary panel, setting an alternate class attribute is also supported (b) and (c).

## 4    Basic Usage

The design considerations detailed in Section 3 are implemented in this first release of DataLearner. This section will now briefly outline their usage.

### 4.1    Loading a dataset

Upon installation or launching the app, the 'Load' screen shown in Fig. 1a greets the user. Pressing the 'Load Weka-format File' button at the bottom of the screen opens up Android's storage access framework (SAF). This controls DataLearner's file access to internal storage (embedded flash and microSD) for loading suitable ARFF files. The user taps the dataset file of interest, as shown in Fig. 2a, and it automatically loads into DataLearner. Once the file is selected and loaded, details of the training dataset appear in the Load screen's summary panel, shown in Figs. 2b and 2c. This panel is scrollable vertically and horizontally, allowing all details to be seen.

Below the summary panel is an option for selecting an alternate class attribute. By default, the class attribute is assumed to be the last attribute listed and the 'Set alternate class attribute' drop-down selection box is set to this attribute index. However, it can be altered by selecting any attribute index from the drop-down list. This setting is only used for classification model building, however, changing the class attribute may aid in further dataset knowledge discovery.

While DataLearner does not presently include functionality for creating a training dataset internally, the ARFF file format is a simple modified CSV (comma-separated variable) text file. This format can be created with any text editor available to Android, or on an external device. Details of the ARFF format are available in [28, 29].

### 4.2   Selecting an algorithm

With a training dataset loaded, the user either swipes left or taps the 'Select' tab at the top of the app display to bring up the algorithm selection fragment as shown in Fig. 3. This menu is divided into two main groups – classification algorithms grouped into six subcategories, plus the clustering and association rule mining algorithms beneath. Only one algorithm can be selected at any one time and this is done by tapping on a subgroup and selecting an algorithm from the drop-down selection.

### 4.3   Running a classification algorithm

Once the training dataset and algorithm have been selected, the user again swipes left or taps the 'Run' tab on the display as shown in Fig. 4a. Pressing the Run button begins the model training process and the Run button toggles into a 'Stop' button. Upon completion, summary results are displayed in the results panel shown in Fig. 4c. In addition, run times for build and ten-fold cross-validation processes are displayed in the scrollable status panel at the bottom of the results screen.

If a clustering or association rule algorithm is selected, DataLearner presents a customised Run tab window, with the results viewable in a large-area text table, as shown in Fig. 5. Again, the screen is fully-scrollable to enable access to all details.

### 4.4   Details and confusion matrix

After completion of the classification build and cross-validation process, the 'View Details and Confusion Matrix' button is enabled. Pressing this button opens a detailed window revealing the accuracy by class, confusion matrix of the model built and classifier output (as appropriate), as shown in Fig. 6. The window is fully-scrollable and can be viewed in portrait (vertical) or landscape (horizontal) mode. Pressing the device's back arrow will return the user to the main DataLearner application screens.

## 5   Accuracy, Speed and Energy Consumption

Smartphones and tablets do not feature the same processor architecture as more traditional desktop and laptop computers. However, this does not imply that mobile devices cannot execute data mining and machine-learning algorithms with the same accuracy. Tests carried out for this paper comparing the accuracy of a number of classification algorithms and datasets on a quad-core Intel desktop PC and a 2017-era

Motorola Moto G5 smartphone using the DataLearner app show this to be the case. Details of the five datasets tested are shown in Table 1. All datasets used in testing are publicly available from the UCI Machine Learning Repository [30].

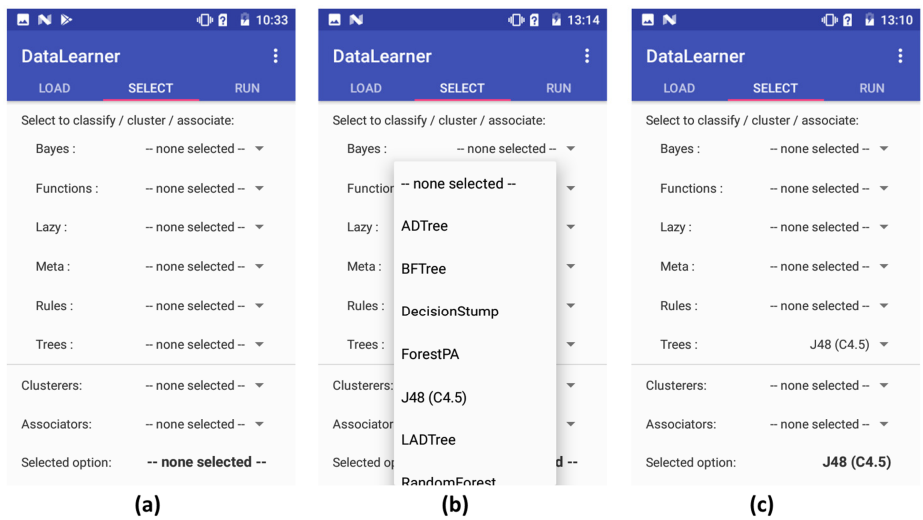

**Fig. 3.** The Select tab provides access to 40 algorithms (a), these can be selected by tapping any drop-down menu (b) and only one algorithm at a time can be selected (c).

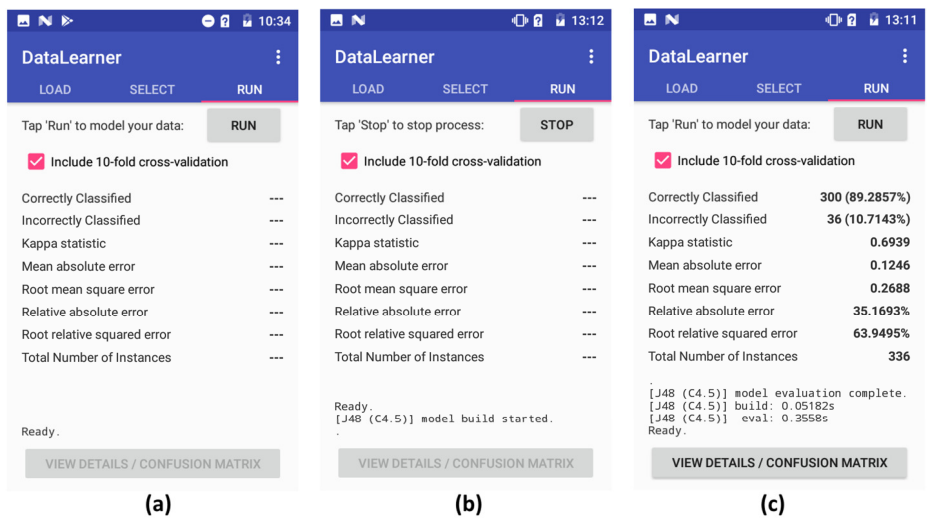

**Fig. 4.** Press the Run button to begin model training (a), use the Stop button to halt training (b), read the results in the summary area when completed (c).

**Table 1:** Publicly-available training datasets used in testing

| Dataset | Instances | Attributes | Attribute Type | Missing Values |
|---|---|---|---|---|
| Car Eval | 1728 | 6 | Categorical | No |
| Ecoli | 336 | 8 | Numeric | No |
| Mushroom | 8124 | 22 | Categorical | Yes |
| Soybean | 683 | 36 | Categorical | No |
| Thyroid | 7200 | 21 | Mixed | No |

**Table 2.** Classification accuracy results comparing a quad-core Intel Core i5-2300 PC using Weka 3.6.15 with a Motorola Moto G5 phone running DataLearner (scores as percentages).

| | 3.1GHz Intel Core i5-2300 PC (running Weka 3.6.15) | | | Motorola Moto G5 smartphone (running DataLearner) | | |
|---|---|---|---|---|---|---|
| **Dataset** | **J48 (C4.5)** | **NaiveBayes** | **REPTree** | **J48 (C4.5)** | **NaiveBayes** | **REPTree** |
| Car Eval | 92.3611 | 85.5324 | 87.6736 | 92.3611 | 85.5324 | 87.6736 |
| Ecoli | 89.2857 | 85.4167 | 90.1786 | 89.2857 | 85.4167 | 90.1786 |
| Mushroom | 100 | 95.8272 | 99.9631 | 100 | 95.8272 | 99.9631 |
| Soybean | 91.5081 | 92.9722 | 84.7731 | 91.5081 | 92.9722 | 84.7731 |
| Thyroid | 99.5758 | 95.281 | 99.5758 | 99.5758 | 95.281 | 99.5758 |

**Table 3.** Classification model build times comparing a quad-core Intel Core i5-2300 PC using Weka 3.6.15 with a Motorola Moto G5 smartphone running DataLearner (scores in seconds).

| | 3.1GHz Intel Core i5-2300 PC (running Weka 3.6.15) | | | Motorola Moto G5 smartphone (running DataLearner) | | |
|---|---|---|---|---|---|---|
| **Dataset** | **J48 (C4.5)** | **NaiveBayes** | **REPTree** | **J48 (C4.5)** | **NaiveBayes** | **REPTree** |
| Car Eval | <0.001 | <0.001 | <0.001 | 0.0479 | 0.0054 | 0.0349 |
| Ecoli | <0.001 | <0.001 | <0.001 | 0.0115 | 0.0074 | 0.0169 |
| Mushroom | 0.01 | <0.001 | 0.03 | 0.2077 | 0.0554 | 0.3812 |
| Soybean | 0.01 | <0.001 | 0.01 | 0.1187 | 0.0079 | 0.1072 |
| Thyroid | 0.03 | 0.01 | 0.02 | 0.2770 | 0.0814 | 0.2476 |

The results of accuracy comparisons are shown in Table 2. Since both platforms ran essentially the same algorithm source code, the classification accuracy results of the two computing devices were identical, as expected. By contrast, smartphones are known not to have the same processor performance levels as desktop and laptop computers. However, as mobile processor development continues to improve, smartphone performance also continues to gain pace. To understand the performance differences, the same combination of algorithms, datasets and devices were again tested for model build times, the results shown in Table 3.

Overall, DataLearner running on the Motorola Moto G5 smartphone delivered approximately one-tenth of the desktop PC's processing speed. However, it must be noted that DataLearner's Weka 3.6.15 core engine only utilised one of the phone's four main processing cores. Nevertheless, all algorithm/dataset combination model builds were completed within half-a-second. Moreover, adding support for multi-core processing into the algorithms should improve processing times considerably.

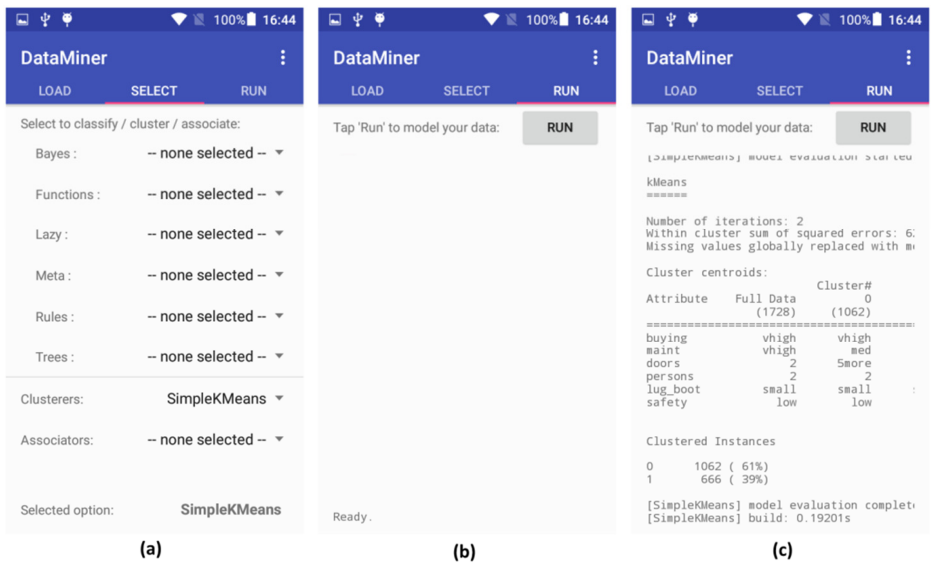

**Fig. 5.** Selecting a clustering or association mining algorithm (a) modifies the Run panel (b) with results shown in a fully-scrollable output (c).

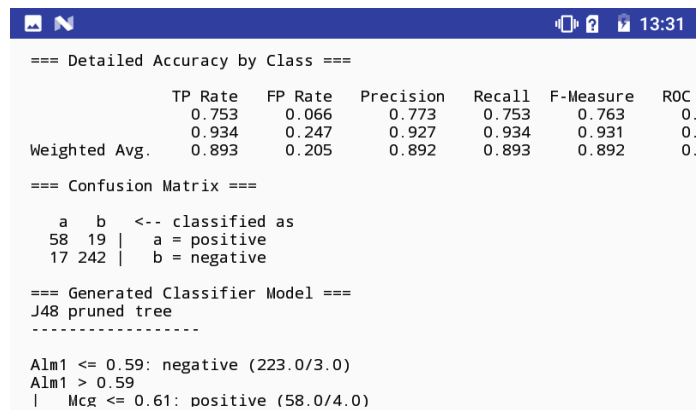

**Fig. 6.** The 'View Details & Confusion Matrix' button launches a full-screen scrollable view of accuracy by class, confusion matrix and classifier output of the new model.

Additionally, testing of DataLearner's energy consumption was conducted, with the J48 (C4.5) algorithm temporarily set to continuously build models from the Thyroid dataset over a 10% range of battery capacity on the Moto G5 phone. This combination completed 7,052 builds within that 10% range. This equates to a single build of this combination using less than 0.000015% of the device's battery capacity.

1212

## 6  Discussion & further research

DataLearner's ability to match PC levels of accuracy on smartphone hardware shows that mobile devices present no impediment when it comes to data mining model training. Moreover, we believe DataLearner can become a useful additional learning tool for where a desktop or laptop computer is inconvenient or unavailable. For training datasets with dimensions similar to those presented in Section 5, the inherently reduced levels of processing speed provided by smartphone-grade hardware should not provide a hindrance. Furthermore, having a broad-based data mining and knowledge discovery tool in a smartphone should provide ample compensation. Add in the continuing improvements in smartphone processing power and this should allow for progressively larger datasets to be processed into the future. Nevertheless, the use of only a single processor core by all algorithms in DataLearner is a limitation that provides opportunities for further research. More recent versions of Weka are known to offer multi-core support in some algorithms (including RandomForest and Bagging).

However, as previously noted, versions of Weka beyond 3.7 feature a user interface based on Java's Swing/Abstract Window Toolkit (AWT), which is not fully supported by Android. Thus, areas for further improvement of DataLearner include the development of multi-threaded algorithms to take advantage of multi-core processors inside modern smartphones, as well as development of more efficient algorithms that can achieve faster processing speeds within existing mobile devices. Other features being considered include a dataset editor to allow users to change individual attribute values within a training dataset. Support for standard .CSV files is also being investigated, along with the possibility of an iOS version for Apple devices. Finally, finding a way to update to newer version 3.8 and 3.9 releases of Weka is also a priority.

## 7  Conclusion

DataLearner turns any smartphone or tablet with Android 4.4 or later into a portable, self-contained, data mining and knowledge discovery tool. It incorporates the open-source Weka data mining engine with minimal modifications to work on Android and provides 40 classification, clustering and association rule mining algorithms.

This paper has discussed the limited research into and examples of general-purpose data mining apps available for mobile devices, as well as the current lack of data mining software available on Google Play. It has detailed the techniques and methods used to incorporate the Weka run-time engine into the Android development platform. It has outlined the implementation and usage of the DataLearner application. Further, performance testing has shown that a smartphone running DataLearner delivers identical levels of accuracy as traditional PCs and despite only using single-threaded code, processing speed is satisfactory and continues to improve. Initial energy consumption testing indicates building a single data mining model on a smartphone using DataLearner has negligible effect on device battery life. Nevertheless, DataLearner has areas for improvement, including multi-core support for algorithms and a dataset editor.

### 7.1 Availability on Google Play, GitHub and YouTube

To our knowledge, DataLearner is the only app of its type currently available on the Google Play store. As outlined in section 1.1, the DataLearner platform consists of four elements – this paper, the app available on Google Play, the GPL3-licensed source code on GitHub, plus a short video on YouTube.

If smartphones and tablets are the ultimate 'personal' computers, they provide the ultimate platform to explore knowledge discovery and data mining in a range of applications, including education, remote-location and personalised e-health. Moreover, having the ability to perform data mining and knowledge discovery in a pocketable device without the need for network connectivity or mains power should continue to open up a broad array of development opportunities.

### Acknowledgments

This research is supported by an Australian Government Research Training Program (RTP) scholarship.

### References

[1] University of Waikato. Weka 3: Data Mining Software in Java. http://www.cs.waikato.ac.nz/ml/weka/, last accessed 2019, 18 May.
[2] n.d. RStudio, open-source and enterprise-ready professional software for R. https://www.rstudio.com/, last accessed 2019, 28 May.
[3] Holmes, G., A. Donkin, and I.H. Witten: Weka: A machine learning workbench. In:  (1994).
[4] Srinivasan, V., S. Moghaddam, A. Mukherji, K.K. Rachuri, C. Xu, and E.M. Tapia: Mobileminer: Mining your frequent patterns on your phone. In: Proceedings of the 2014 ACM International Joint Conference on Pervasive and Ubiquitous Computing. ACM:  (2014).
[5] Nath, S.: ACE: exploiting correlation for energy-efficient and continuous context sensing. In: Proceedings of the 10th international conference on Mobile systems, applications, and services. ACM:  (2012).
[6] n.d. TensorFlow For Mobile & IoT - Overview. https://www.tensorflow.org/mobile/, last accessed 2019, 8 Jun.
[7] Hall, M., E. Frank, G. Holmes, B. Pfahringer, P. Reutemann, and I.H. Witten: The WEKA data mining software: an update. In: ACM SIGKDD explorations newsletter. **11**(1): p. 10-18 (2009).
[8] Banos, O., R. Garcia, J.A. Holgado-Terriza, M. Damas, H. Pomares, I. Rojas, A. Saez, and C. Villalonga: mHealthDroid: a novel framework for agile development of mobile health applications. In: International workshop on ambient assisted living. Springer:  (2014).
[9] rjmarsan. Weka-for-Android. https://github.com/rjmarsan/Weka-for-Android, last accessed 2019, 19 May.

[10] BinDhim, N.F., A.M. Shaman, L. Trevena, M.H. Basyouni, L.G. Pont, and T.M. Alhawassi: Depression screening via a smartphone app: cross-country user characteristics and feasibility. In: Journal of the American Medical Informatics Association. **22**(1): p. 29-34 (2015).

[11] Jeong, T., D. Klabjan, and J. Starren: Predictive Analytics Using Smartphone Sensors for Depressive Episodes. In: arXiv preprint arXiv:1603.07692. (2016).

[12] Allouch, A., A. Koubâa, T. Abbes, and A. Ammar: RoadSense: Smartphone Application to Estimate Road Conditions using Accelerometer and Gyroscope. In: IEEE Sensors Journal. **17**(13): p. 4231 (2017).

[13] Yates, D., M.Z. Islam, and J. Gao: Implementation and Performance Analysis of Data-Mining Classification Algorithms on Smartphones. In: Australasian Conference on Data Mining. Springer: (2018).

[14] Dou, X. and S.S. Sundar: Power of the swipe: Why mobile websites should add horizontal swiping to tapping, clicking, and scrolling interaction techniques. In: International Journal of Human-Computer Interaction. **32**(4): p. 352-362 (2016).

[15] Package weka.classifiers.bayes. http://weka.sourceforge.net/doc.dev/weka/classifiers/bayes/package-summary.html, last accessed 2019, 29 May.

[16] Package weka.classifiers.functions. http://weka.sourceforge.net/doc.dev/weka/classifiers/functions/package-summary.html, last accessed 2019, 29 May.

[17] Package weka.classifiers.lazy. http://weka.sourceforge.net/doc.dev/weka/classifiers/lazy/package-summary.html, last accessed 2019, 8 June.

[18] Package weka.classifiers.meta. http://weka.sourceforge.net/doc.dev/weka/classifiers/meta/package-summary.html, last accessed 2019, 29 May.

[19] Package weka.classifiers.rules. http://weka.sourceforge.net/doc.dev/weka/classifiers/rules/package-summary.html, last accessed 2019, 29 May.

[20] Package weka.classifiers.trees. http://weka.sourceforge.net/doc.dev/weka/classifiers/trees/package-summary.html, last accessed 2019, 29 May.

[21] Package weka.clusterers. http://weka.sourceforge.net/doc.dev/weka/clusterers/package-summary.html, last accessed 2019, 29 May.

[22] Package weka.associations. http://weka.sourceforge.net/doc.dev/weka/associations/package-summary.html, last accessed 2019, 29 May.

[23] Islam, Z. and H. Giggins: Knowledge discovery through SysFor: a systematically developed forest of multiple decision trees. In: Proceedings of the Ninth Australasian Data Mining Conference-Volume 121. Australian Computer Society, Inc.: (2011).

[24] Adnan, M.N. and M.Z. Islam: Forest PA: Constructing a decision forest by penalizing attributes used in previous trees. In: Expert Systems with Applications. **89**: p. 389-403 (2017).

[25] Yates, D., M.Z. Islam, and J. Gao: SPAARC: A Fast Decision Tree Algorithm. In: Australasian Conference on Data Mining. Springer: (2018).

[26] Breiman, L., J. Friedman, C.J. Stone, and R.A. Olshen: Classification and regression trees. CRC press (1984).

[27] Distribution dashboard. https://developer.android.com/about/dashboards/, last accessed 2019, 17 May.

[28] Witten, I.H., E. Frank, and M.A. Hall: Data Mining: Practical machine learning tools and techniques. Morgan Kaufmann (2011).

[29] Weka Wiki - ARFF (stable version). https://waikato.github.io/weka-wiki/arff_stable/, last accessed 2019, 8 June.

[30] Dheeru, D., Karra Taniskidou, E. UCI Machine Learning Repository. https://archive.ics.uci.edu/ml/datasets.html, last accessed 2019, 12 May.